\documentclass{article}
\usepackage{graphicx}

\usepackage{authblk}

\begin{document}

\title{Intelligence plays dice: Stochasticity is essential for machine learning}

\author[1]{Mert R. Sabuncu}
\affil[1]{School of Electrical and Computer Engineering\\ 
Meinig School of Biomedical Engineering\\ 
	Cornell University, Ithaca, NY}

\maketitle




\begin{abstract}
Many fields view stochasticity as a way to gain computational efficiency, while often having to trade off accuracy. In this perspective article, we argue that stochasticity plays a fundamentally different role in machine learning (ML) and is likely a critical ingredient of intelligent systems. As we review the ML literature, we notice that stochasticity features in many ML methods, affording them robustness, generalizability, and calibration. We also note that randomness seems to be prominent in biological intelligence, from the spiking patterns of individual neurons to the complex behavior of animals. We conclude with a discussion of how we believe stochasticity might shape the future of ML.  

\end{abstract}




\section{Introduction}
\label{sec:introduction}
Classical problems in engineering and computer science demand precision and reliability. When solving an equation, using the result to encode a message, transmitting the coded message to another device, decoding the message at the other end, saving the message onto a hard drive, or using it to create a visual rendering; inaccuracies are often the system`s enemy and have to be fought against. Furthermore, if and when any of these computational operations is repeated, we expect the results to be unchanged. We view an unrepeatable result as a sign of a ``bug'' that either has to be fixed, tamed, or at least well understood and tolerated. Reduced precision and reliability is often considered as a price in the tradeoff with computational efficiency. 
The central thesis of this perspective article is that for machine learning (ML) specifically, and artificial intelligence (AI) more generally, probabilistic operations are fundamentally important building blocks, which the field is growing to rely on. We anticipate that stochasticity will therefore feature more prominently, and as a fundamental principle, in the future of machine intelligence.

In the remainder of this article, we refer to a computation that produces a result with a random (i.e., not predetermined and exact) component as stochastic. That is, each time a stochastic computation is executed, the result will, in general, fluctuate around an average (or mean) value. This random fluctuation, importantly, can be statistically independent from the mean value and can be well-characterized, for example, by its probability distribution. Note that, in algorithmically introduced stochasticity, we rely on a pseudo-random number generator, which, of course, can be seeded deterministically, if, for instance, the results need to be repeated. Stochastic computations form the backbone of various approximate optimization techniques, where the objective is to achieve a solution that is as accurate as possible within a computational budget. This general usage of stochasticity, in our view, is in stark contrast with its primary utility in artificial intelligence, where stochasticity can promote adaptivity, flexibility, and robustness. 

\begin{figure}[t!]
\begin{centering}
\includegraphics[width=0.7\textwidth]{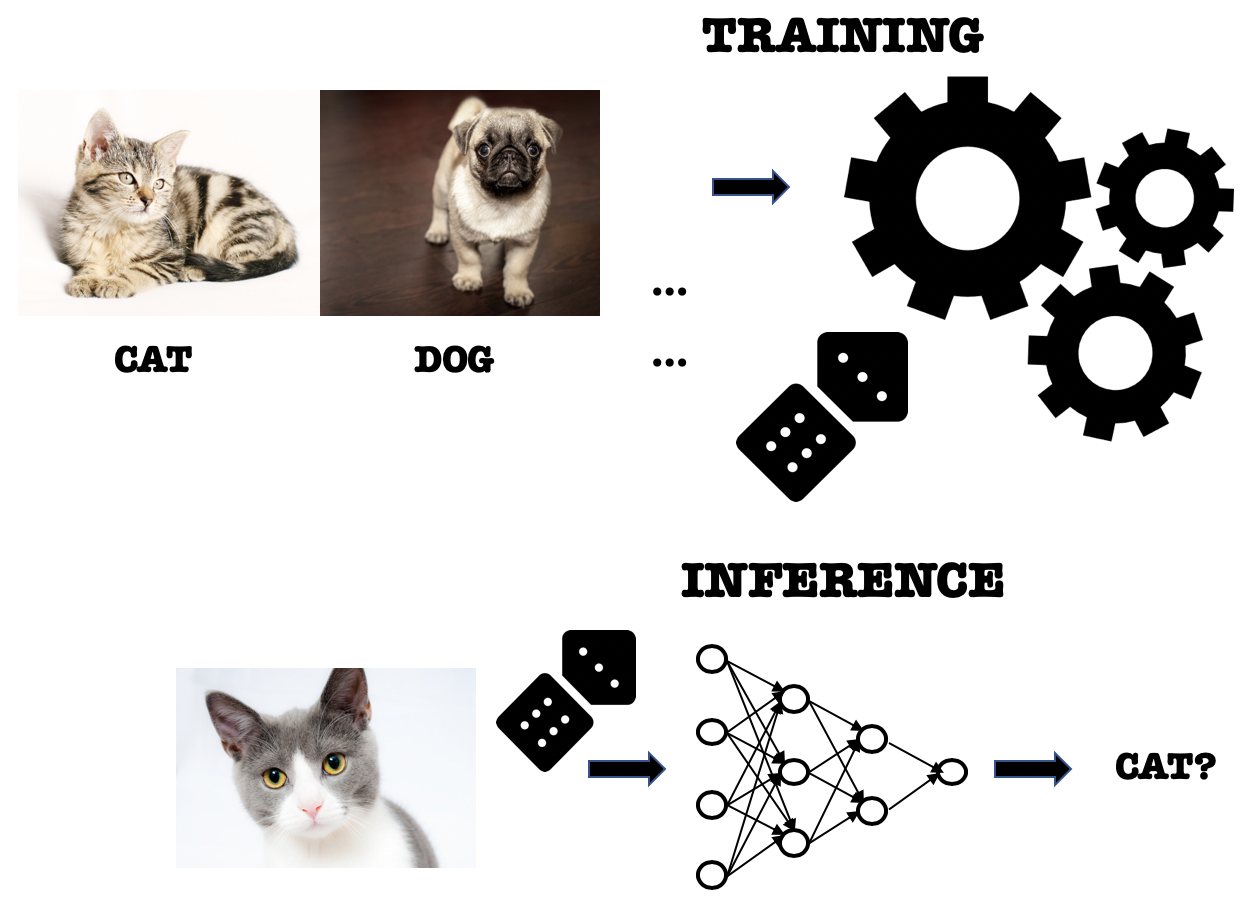}
\caption{Intelligence plays dice. Most modern ML algorithms involve stochasticity, both during training (e.g., stochastic gradient descent) and test-time inference (e.g., Monte Carlo methods).} 
\label{fig:1}
\end{centering}
\end{figure}

\subsection{What makes machine learning unique?}

Machine learning stands apart from most other computational areas, mainly because of the degree of intrinsic uncertainty associated with the domain. For the wide spectrum of real-world settings that ML deals with, the past and the future can look wildly different. This variation is what makes ML problems interesting and challenging. This is in contradistinction to conventional problems in areas like communication or software engineering, where uncertainty is treated as nuisance and can be modeled as noise, which is often theoretically or empirically characterized. The focus, therefore, of conventional computational fields is on solving the problem at hand as accurately as possible, while controlling for the effects of noise. In fact, in many areas, we have become so good at controlling noise that it is not a significant part of the design consideration. For instance, today, software developers worry little about errors due to imperfections in hardware. 

In ML, the large degree of uncertainty means we have a strong distinction between the data used to develop a model (often called training data) and the data used to objectively assess performance (the validation and test datasets), sometimes under different conditions. Most ML models exhibit poorer performance on independent data than on training data - a phenomenon called overfitting. Overfitting, or the generalization gap (performance difference between training and test data), is one of the core challenges of ML.

Overfitting has been the focus of ML research since the inception of the field, and researchers have developed a growing set of techniques to combat overfitting. These techniques typically fall under the following three categories:

\begin{enumerate}

\item Constraining model capacity (or expressivity of the hypothesis space): Classically, the Vapnik-Chervonenkis (VC) theory provides a theoretical underpinning for this approach~\cite{vapnik2013nature}. In recent years, modern manifestations of this idea include the adoption of convolutional and other types of neural network architectures that implement symmetry properties~\cite{cohen2016group}. The main drawback of model capacity restriction is that many real-world problems do not exhibit symmetry properties that can be easily exploited in this manner. For example, in an image classification problem, the labels of the object in a scene might be invariant to changes in lighting or camera position and angle, but such invariances are difficult to bake into a model.

\item Regularization: Instead of applying strict restrictions on model capacity that can bound performance, an alternative strategy is to implement softer constraints, often inspired by inductive biases about the problem. Popular examples of regularization include LASSO~\cite{tibshirani1996regression} and Dropout~\cite{srivastava2014dropout}. Regularization is arguably more flexible and powerful than model capacity restriction and today, it is probably the most widely used framework for combating overfitting. 

\item Ensembling: Another way to mitigate overfitting involves averaging over multiple models, instead of relying on one. Example methods of ensembling include boosting~\cite{freund1997decision}, bootstrap aggregating or bagging~\cite{breiman1996bagging}, and Bayesian averaging~\cite{mackay2003information}. The widely used random forests algorithm~\cite{breiman2001random} is a canonical example of the idea of ensembling. 
\end{enumerate}

In the following sections, we will discuss the growing role of stochasticity in ML, in part, by referring to the above mentioned techniques and the overfitting challenge. But first, let us take a small detour and turn our attention to neuroscience, which has provided inspiration for some of the most successful machine learning algorithms to date. It is definitely true that machine intelligence does not require a grounding in neuroscience. However, we believe that if there is a fundamental principle that underlies intelligence, learning, and adaptive behavior, echoes of it should be present in biology. As we briefly discuss below, stochasticity seems to be heavily featured in neuroscience.

\begin{figure}[t!]
\begin{centering}
\includegraphics[width=0.7\textwidth]{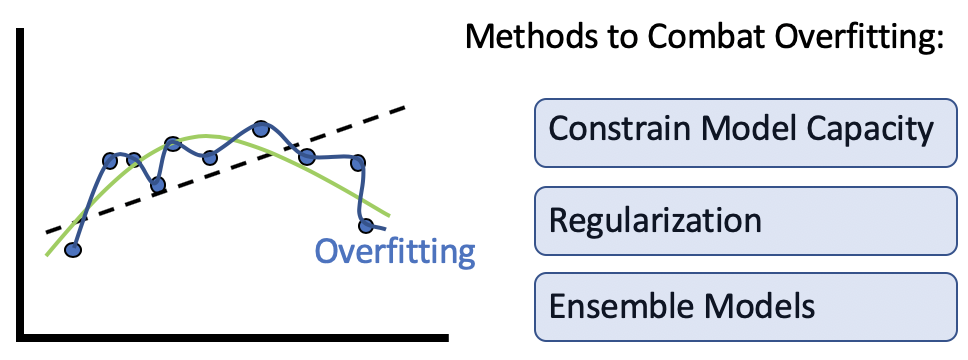}
\caption{Overfitting is a core challenge of ML} 
\label{fig:2}
\end{centering}
\end{figure}

\subsection{Observations from neuroscience}
There is increasing evidence from biological studies that the structure and function of the neural substrate seems to be shaped substantially by stochastic elements. For instance, the firing of an individual neuron is not deterministic and reliable~\cite{shadlen1994noise,czanner2015measuring}. Experimental data suggest that synaptic transmission is mediated by the stochastic release of neurotransmitters, and membrane potentials oscillate due to random variations in ion channels~\cite{schneidman1998ion}. 
The probabilistic spiking of neurons has been argued to be useful in memory, decision making, and attention~\cite{deco2009stochastic}.
Random fluctuations also seem to play an important role in the cellular-level developmental fate of a neuron, which determines its projections and thus the wiring architecture of the brain~\cite{honegger2018stochasticity}. Furthermore, one of the main roles of some neural circuit components seems to be to promote stochasticity in brain function. For example, silencing some neurons in the fruit fly central complex modifies the variability of locomotor behavior without altering the population average~\cite{honegger2018stochasticity}. 

Perception exhibits stochastic features too. As an illustration, it is now well-established that binocular rivalry~\cite{Fox1967}, i.e., the alternation in the perception of different images presented dichoptically, is a stochastic process. Similarly, stochastic switching is a well documented phenomenon in auditory bistable perception and taste processing~\cite{Gigante,Miller}.

At a higher level, animal behaviors can display strong stochasticity also. For instance, non-human primates produce highly randomized choice behavior, especially in the context of competitive interactions with other individuals~\cite{Soltani}. In another example, individual locusts appear to increase the randomness of their movements to maintain swarm alignment in sudden coherent changes in direction of the group movement~\cite{Yates}. Genetically identical organisms reared under identical conditions can also exhibit widely variable behaviors~\cite{vogt2008production}. For example, pea aphids display stochastic variability in their escape responses. When a threatening object appears, aphids feeding on vegetation react randomly: some individuals jump away quickly, while others remain feeding~\cite{schuett2015life,honegger2018stochasticity}. Recently, it has also been postulated that dreams in animals might have the evolutionary purpose of combating overfitting to past experience, which is achieved by injecting stochasticity into learning and thus allowing departures from the statistical patterns of an animal's daily life~\cite{hoel2020overfitted}. Thus, dreams can assist with generalization and therefore future performance.

\begin{figure}[t!]
\begin{centering}
\includegraphics[width=0.7\textwidth]{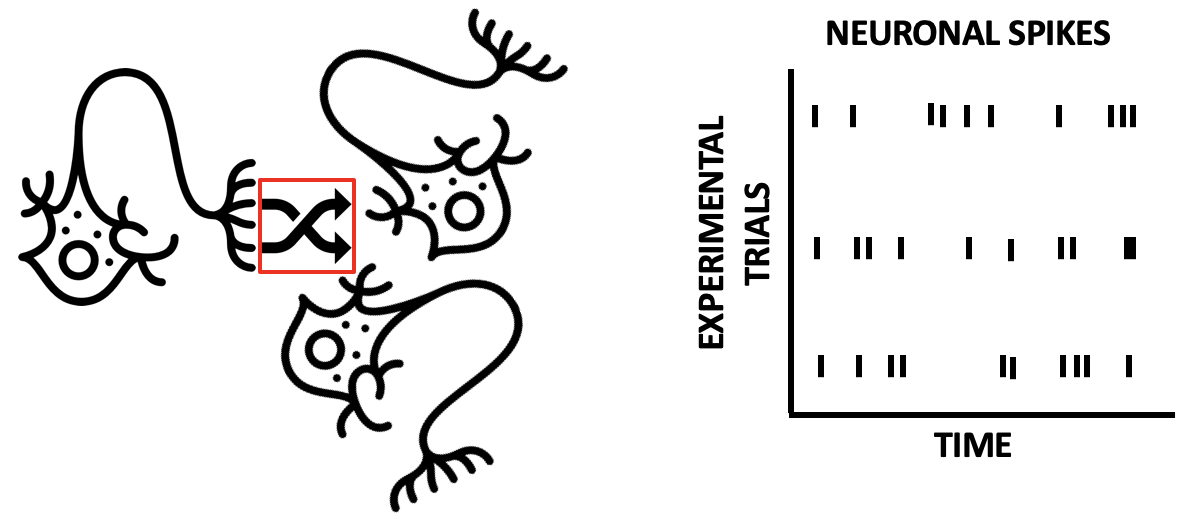}
\caption{Stochasticity is prominent in biological intelligence. Random noise plays a role in establishing synaptic connections between neurons.The firing patterns of individual neurons are not deterministic. } 
\label{fig:3}
\end{centering}
\end{figure}

\section{A review of stochasticity in machine learning}

Now, let us return to the field of machine learning.
Many of the classical ML algorithms, such as support vector machines~\cite{cortes1995support}, hand engineered decision trees~\cite{quinlan1987simplifying}, regularized regression~\cite{tibshirani1996regression}, and Gaussian processes~\cite{rasmussen2003gaussian} did not heavily rely on randomization.
While stochasticity first made a real appearance in ML in the early 90's, it is not until recently that it became ubiquitous.
In fact, today virtually every single modern ML algorithm that is implemented features stochasticity at its core.
In this section, we provide an overview of some of the most influential ideas that build on randomization, mostly restricting ourselves to the context of supervised learning, where the objective is to fit a model that maps some input data to an output label, based on minimizing a loss function on a given set of labeled training data.

\begin{figure}[t!]
\begin{centering}
\includegraphics[width=0.7\textwidth]{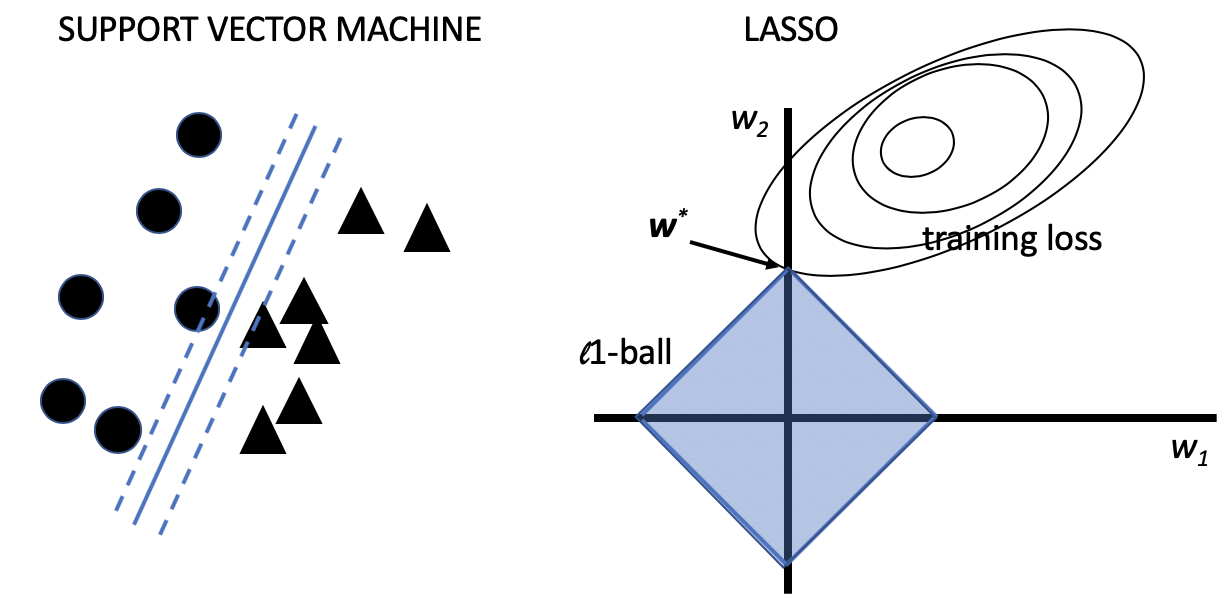}
\caption{Stochasticity was not a mainstay feature of many classical ML algorithms, such as Support Vector Machines~\cite{cortes1995support} and LASSO~\cite{tibshirani1996regression}.} 
\label{fig:2}
\end{centering}
\end{figure}

\subsection{Stochasticity in data generation}
One of the first steps where noise injection is widely used is during the creation of the data a model is trained on. For example, an approach that was originally developed in statistics and later adopted for machine learning is bootstrapping, where random training datasets are generated by sampling with replacement. This allows one to train different models, which can be combined in an ensemble - a method invented by Leo Breiman who called it bootstrap aggregating, or bagging~\cite{breiman1996bagging}. Another technique that is widely used to exploit symmetries in the problem is data augmentation~\cite{perez2017effectiveness}, where random transformations are applied to the original data that either conserve labels or alter them in a predictable manner. A popular example is random rotations and other geometric transformations used in computer vision problems, where a label of an object in an image (say, a cat) is unaltered by rotation. One can also add noise to the input data during training in order to improve robustness. For instance, a widely used method in computer vision involves randomly erasing a small rectangular patch in an image~\cite{zhong2020random}. An alternative strategy is to corrupt the labels during training, where a small fraction of randomly selected training labels can be randomly swapped~\cite{xie2016disturblabel} or probabilistically smoothed. Mixup~\cite{zhang2017mixup} is another method popular in deep learning, where additional training samples are generated by convexly combining random pairs of samples and their associated labels. Finally, we have domain randomization~\cite{tobin2017domain} - a technique that uses randomly chosen settings for a data synthesizer to create a wide range of training samples with a substantial amount of variability. This approach has been shown to improve real-world performance drop.

\subsection{Stochastic modeling}
 
In supervised ML, a model describes an idealized relationship between the input data and the label. Traditionally, the choice of a model is based on an understanding of the real-world mechanisms that generate the data and/or mathematical or computational convenience. Once a model decision has been made, training proceeds via running an optimization algorithm on some labeled data. Instead of committing to a model choice, however, an alternative is to consider a collection of random models. One of the earliest proponents of this idea was Eugene Kleinberg, who formally studied a form of stochastic modeling, which gave rise to the popular random forest algorithm~\cite{kleinberg2000algorithmic,breiman2001random}. In this framework, each model is restricted to a random subspace of the input domain. For example, a random subset of variables is often used to build a so-called weak model in random forests.  More recently, stochasticity has been featured in neural network architecture design too. For instance, in deep networks with stochastic depth, the number of layers are chosen randomly~\cite{huang2016deep}. An alternative approach is to use a random search algorithm to optimize over the space of neural network architectures~\cite{xie2018snas,li2019random}. 

\subsection{Stochastic optimization}

Once the training dataset and model have been established, the next phase in the ML workflow entails running an optimization procedure. In optimization, the objective is to find the best solution (in the case of ML, the set of model parameters that minimize a loss function over a training set), often constrained by a computational budget. Global optima are usually hard to determine outside of the convex setting and/or in high dimensional data, which virtually all interesting ML problems suffer from. Stochastic techniques have been originally employed for ML to achieve computational efficiency and address the challenge of non-convexity. For example, stochastic gradient descent (SGD) was first popularized in ML as a way to efficiently perform approximate optimization over very large scale datasets~\cite{bottou2010large}. SGD and its variants like ADAM~\cite{kingma2014adam} are now the primary workhorses behind most modern ML algorithms. Similarly, multiple random initializations has been widely used in an attempt to overcome the non-convexity challenge. However, recently there has been a deeper appreciation for the stochasticity in these techniques. For instance, SGD has been shown to implement approximate Bayesian inference~\cite{mandt2017stochastic,smith2017bayesian} or act as a regularizer~\cite{zhang2016understanding}, which can explain why it leads to solutions that generalize well. Furthermore, SGD has been shown to be more immune to getting stuck in local optima than its deterministic counterpart~\cite{kleinberg2018alternative}. 
Random initializations have also been employed to create ensembles of models that offer improved performance, as shown in~\cite{lakshminarayanan2017simple}. 
It was recently demonstrated that randomly-initialized, feed-forward networks include subnetworks (so-called ``winning tickets'') that can achieve test accuracy comparable to the original network with similar amount of training~\cite{frankle2018lottery}. 
Another related idea is the so-called Extreme Learning Machine (ELM), where many of the model parameters (e.g., all weights of the hidden neurons in a neural network) are randomized and not optimized~\cite{huang2006extreme}. In ELMs, optimization is focused on a small set of the parameters, such as the weights of the output layer. This strategy has been shown to significantly boost the performance and/or efficiency of neural networks. Finally, non-gradient descent based optimization techniques, such as genetic algorithms~\cite{goldberg1988genetic} and simulated annealing~\cite{aarts1988simulated}, also heavily feature randomness and promise to shape the future of machine learning, as the field evolves beyond the current era of back-propagation based deep learning. It is important to note that gradient-free stochastic optimization methods are already in widespread use for efficient policy search in reinforcement learning, e.g.~\cite{mania2018simple}.

\subsection{Random noise injection} 

As the field of ML has stopped regarding stochasticity as an approximation strategy for computational efficiency, but rather as an approach that yields robust and accurate models, new ways of introducing randomness have also been proposed. One of the most successful examples of this is dropout~\cite{srivastava2014dropout}, where units in a neural network are randomly erased during training. Dropout was originally proposed as a simple regularization technique, but has also been shown to enable ensembling at test time~\cite{gal}. Dropout can be generalized, as in structured dropout~\cite{zhang2019confidence}, where groups of neurons are erased, or with noise injection, where instead of erasing neurons, their activations are randomly perturbed~\cite{neklyudov2017structured}.

In the context of unsupervised learning and generative modeling, random noise can play a critical role too. For instance, in the popular variational auto-encoder (VAE)~\cite{vae}, a stochastic layer offers a way to account for uncertainty and ensure statistical regularity in representation learning. It also enables the user to generate synthetic data via pushing random noise instantiations through the decoder. Generative Adversarial Networks (GANs)~\cite{gan} and Normalizing Flows~\cite{flow} also rely on the idea of mapping random noise to realistic data, for the purposes of synthesis and density estimation.

\begin{figure}[t!]
\begin{centering}
\includegraphics[width=0.7\textwidth]{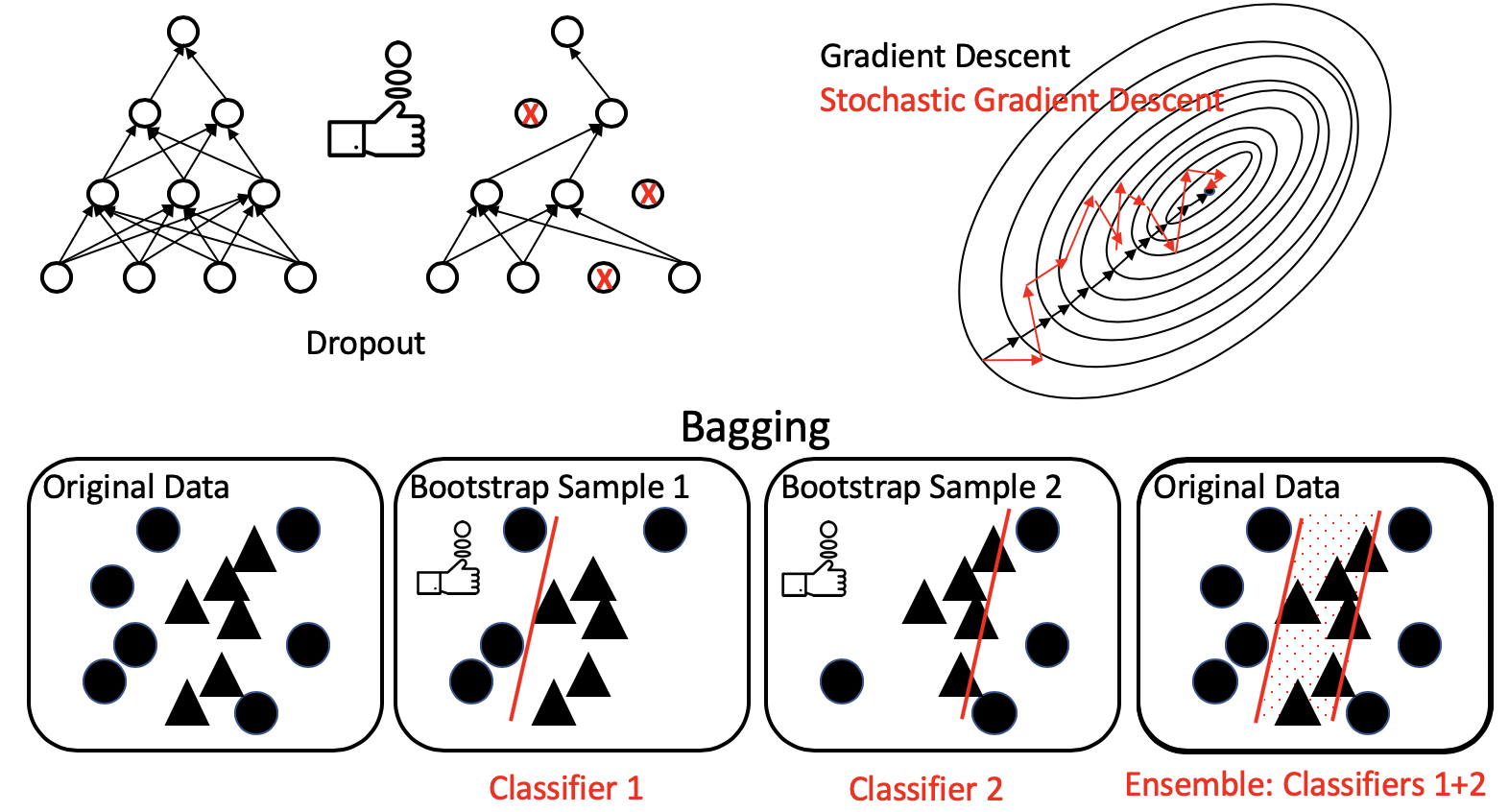}
\caption{Some example widely-used methods in ML that are built on stochasticity: Dropout~\cite{srivastava2014dropout}, Stochastic Gradient Descent~\cite{bottou2010large}, and Bagging~\cite{breiman1996bagging}.} 
\label{fig:4}
\end{centering}
\end{figure}

\subsection{Monte Carlo inference}

Once training is complete, for a given input, most ML models provide a deterministic output (which can be a probability vector for classification problems) during test time (also called inference). A notable exception is Monte Carlo techniques, which involve stochasticity in the evaluation of the model. In this framework, each instantiation of model evaluation (e.g., forward pass through the neural network) can be viewed as a sample from some posterior distribution. Test time dropout, that is applying dropout during test time inference, has shown to implement Monte Carlo sampling~\cite{gal}. Monte Carlo inference can also be implemented more directly via the use of stochastic neurons - an idea pioneered by Radford Neal in the early 90's~\cite{neal2012}. The popular Restricted Boltzmann Machines also feature Monte Carlo techniques~\cite{nair}. Recently, more efficient strategies have been proposed for stochastic neural networks that are better suited to large-scale problems. These include back-propagation based gradient estimation techniques that use the re-parametrization trick~\cite{vae, gumbel, bengio2013} and extensions of the REINFORCE estimator~\cite{williams}.

\section{How stochasticity helps address current challenges in ML}

The stochastic methods we discussed above have proven to be invaluable in helping with addressing several important challenges in ML. Let us briefly summarize these here. 

\subsection{Generalization}
As we mentioned at the beginning, generalization, that is the performance difference between training and test data, is a core challenge of ML. Stochasticity is a crucial ingredient in several effective regularizers, as nicely exemplified in dropout, data augmentation, or label smoothing, and enables ensembling, as in bagging and Bayesian techniques that feature Monte Carlo inference. 

\subsection{Uncertainty estimation}
In many real-world applications such as healthcare and autonomous navigation, accuracy is not the only concern, rather it is also critical to provide useful estimates of uncertainty around the predictions. In classification, this is often referred to as confidence calibration~\cite{calibration}. Many ML models are notoriously overconfident, providing erroneous classifications with high probabilities. Stochasticity has featured in some recent methods that aim to ameliorate this problem. For example, dropout has been shown to improve calibration in neural networks~\cite{zhang2019confidence}.

\subsection{Domain adaptation and transfer learning}
When an ML model is trained and evaluated, this is usually done for a specific problem setting, frequently called a domain. However, we are often interested in deploying these models in other domains, which can be slightly different than the original setting or even show dramatic changes. In the latter scenario, we need to adapt (or fine-tune) our model, either with labeled (supervised) or unlabeled data (unsupervised). In the former scenario, we might be more concerned with limiting the performance degradation due to domain shift. There are a rapidly growing number of methods to address these challenges and many feature stochasticity at their core. For example, denoising autoencoders that are trained with random corruption of the data, can be used for domain adaptation~\cite{domain}. Similarly, domain randomization is a popular technique for achieving robustness against domain shifts~\cite{tremblay2018training}. Finally, many successful domain adaptation techniques rely on data synthesis and adversarial learning techniques (as in GANs), which build on stochastic elements~\cite{sankaranarayanan,long}. 

\subsection{Security}
Many ML models can be susceptible to adversarial attacks, where often imperceptible levels of perturbations of the input can be designed to substantially degrade performance~\cite{Attack}. This phenomenon raises important security concerns that are now widely studied. The ML toolkit contains a growing arsenal to fight against adversarial attacks and most of these methods contain stochastic elements. For example, an effective strategy for defending against adversarial attacks is stochastic activation pruning (SAP), which can be applied to a pre-trained network~\cite{SAP}. In SAP, a random subset of activations (preferentially pruning those with smaller magnitude) are pruned and surviving neurons are amplified to compensate. It has also been recently demonstrated that neural networks with random weights and biases enjoy adversarial robustness guarantees~\cite{de2020adversarial}. On the attack side, the synthesis of adversarial examples that can confound a classifier often involve a randomized component~\cite{Athalye}. 

\subsection{Computational Efficiency}
Many real-world ML problems require handling large-scale data with limited computational resources. Arguably, stochastic gradient descent (SGD)~\cite{bottou2010large} and its variants such as ADAM~\cite{kingma2014adam} represent the most important breakthroughs that have made learning at large scale feasible. Lately, quantization has been recognized as another effective
strategy to satisfy the memory and energy supply demanded by deep neural network models. In quantization, weights, activations, and gradients are represented with low precision. Many of the popular quantization methods feature stochasticity, as in the random rounding used in BinaryConnect~\cite{Binaryconnect}.

\section{Outlook}

In this Section, let us briefly discuss several challenges of ML that, we believe, need more attention and promise some future breakthroughs, as we are armed with an appreciation for stochasticity.
    
\subsection{Hardware}
The latest phase of ML (dominated by deep learning), was in part facilitated by the co-option of graphics processing units (GPUs), which are optimized for rapid matrix computations. While GPUs have undoubtedly accelerated progress in ML and AI, they are ultimately designed for computational tasks that demand precision and reliability. As we argue in this paper, robust and useful ML models can be built on unreliable or stochastic components. Thus, we believe that the future of ML will be fueled by non-traditional hardware that is low precision, energy efficient, and enables stochastic computing. Some recent neuromorphic paradigms demonstrate the different possibilities in this direction, e.g.~\cite{Yu2013}. 

\subsection{Explainability}
Lately, explainable or interpretable ML has become an area that has attracted a lot of interest~\cite{Doshi-Velez, Lipton2018}. This is because in many real-world applications, maximizing prediction accuracy is not the only goal. It is also important to gain insights into the computed prediction, which might in turn have legal implications, shed light on causal mechanisms, inform downstream decision making, or help establish trust between human users and AI. Stochasticity can be a helpful tool for gaining insights from an ML model. For example, a simple method to interpret a convolutional neural network involves perturbing the input image with random noise~\cite{smilkov2017smoothgrad}. For many of the model interpretation use cases, however, we expect that stochasticity will complicate the problem. Thus, we believe, there is more research to be done in this direction, as stochasticity becomes a mainstay of ML.    

\subsection{Bias and Fairness}
As ML and AI tools are widely used in real world decision making, fairness and bias have become important topics to study~\cite{Mehrabi}. One of the most significant goals in real world deployment is that AI-based decisions do not discriminate toward certain individuals or groups, which forms a strong foundation for fairness. Fairness can be considered as one of several objectives for a ML algorithm, which can be theoretically analyzed using the concept of Pareto optimality. As recently demonstrated, stochasticity can play an important role in a multi-objective optimization framework that aims to compute the Pareto fronts of the accuracy and fairness loss-scape~\cite{Liu2020}.

\subsection{Fundamental Principles of ML}
Much of the developments in ML over the last decade have come on the empirical front, where theory has lagged practice. We still lack widely accepted principles that allow us to understand, characterize, and predict the performance of modern ML models, although several important breakthroughs have been made in this direction. For example, Belkin et al. have recently presented a new perspective that reconciles classical theory that prescribes a tradeoff between bias and variance, with modern practice that relies on over-parameterized models~\cite{Belkin}. This perspective has underscored the importance of stochasticity as a regularizer in modern ML. Another promising direction is Tishby et al.'s Information Bottleneck Principle (IBP)~\cite{tishby2015deep}, which studies the tradeoff between prediction accuracy and information compression to provide generalizable insights into the behavior of deep learning models. It was recently pointed out that stochasticity is necessary for certain networks to exhibit information compression in IBP~\cite{Goldfeld}. Similarly, we believe that as fundamental principles for ML emerge, they will feature stochasticity at their core.

\section{Conclusion}
Many interesting and important aspects of our world are not fully predictable. That is why there is a limit to what our high fidelity systems that are engineered to be reliable and precise, can achieve. Intelligence, i.e., the ability to learn from experience and deal with novel situations, demands flexibility, creativity, and adaptation. In this article, we argued that stochasticity, i.e. randomness, can be an important ingredient for building intelligent systems. There is increasing evidence that stochasticity plays a central role in biological intelligence. Machine learning has also grown to rely on stochasticity in developing useful algorithms. Stochasticity's utility in ML goes beyond computational efficiency. It affords robustness, generalizability and calibration. A renewed appreciation for stochasticity in ML, we believe, will inspire the next generation in AI-centric hardware design. It will also fuel further research into building more secure, robust, unbiased, and interpretable ML solutions. 

\section{Acknowledgements}
This work was supported in part by the NIH under grants R01LM012719, R01AG053949, and in part by the NSF under
grants CAREER 1748377 and NeuroNex 1707312.
I would like to thank Zhilu Zhang and Meenakshi Khosla for their valuable input.
The illustrations include several icons obtained from www.flaticon.com. 

\bibliography{MyReferences.bib}{}

\bibliographystyle{plain}

%


\end{document}